# Directed evolution algorithm drives neural prediction

Yanlin Wang[1,2], Nancy M. Young[3,4,5], and Patrick C. M. Wong[1,2*]
[1] Brain and Mind Institute, The Chinese University of Hong Kong, Hong Kong SAR, China
[2] Department of Linguistics and Modern Languages, The Chinese University of Hong Kong, Hong Kong SAR, China
[3] Division of Otolaryngology, Ann & Robert H. Lurie Children's Hospital of Chicago, Chicago, IL, USA
[4] Department of Otolaryngology–Head and Neck Surgery, Feinberg School of Medicine, Northwestern University, Chicago, IL, USA
[5] Knowles Hearing Center, Department of Communication Sciences and Disorders, Northwestern University, Evanston, IL, USA
** Corresponding author: Patrick C. M. Wong, PhD*
Brain and Mind Institute and Department of Linguistics and Modern Languages, The Chinese University of Hong Kong, Hong Kong SAR, China
*Email: p.wong@cuhk.edu.hk*

**Abstract—**
Neural prediction offers a promising approach to forecasting the individual variability of neurocognitive functions and disorders and providing prognostic indicators for personalized invention. However, it is challenging to translate neural predictive models into medical artificial intelligent applications due to the limitations of domain shift and label scarcity. Here, we propose the directed evolution model (DEM), a novel computational model that mimics the trial-and-error processes of biological directed evolution to approximate optimal solutions for predictive modeling tasks. We demonstrated that the directed evolution algorithm is an effective strategy for uncertainty exploration, enhancing generalization in reinforcement learning. Furthermore, by incorporating replay buffer and continual backpropagate methods into DEM, we provide evidence of achieving better trade-off between exploitation and exploration in continuous learning settings. We conducted experiments on four different datasets for children with cochlear implants whose spoken language developmental outcomes vary considerably on the individual-child level. Preoperative neural MRI data has shown to accurately predict the post-operative outcome of these children within but not across datasets. Our results show that DEM can efficiently improve the performance of cross-domain pre-implantation neural predictions while addressing the challenge of label scarcity in target domain.

## I. INTRODUCTION

Neural prediction—using brain data to predict health and behavioral characteristics such as mental health and developmental status—is an area of increasing interdisciplinary research and emerging clinical applications. This application relies fundamentally on a computational framework that leverages advances in machine learning and neuroscience to translate neural data into clinically meaningful forecasts. With the rapid expansion of neuroscience and the advance of machine learning algorithms, there is a considerable need to deliver neural predictive models with individual-level precision to inform clinical decision-making. However, AI applications designed for clinical use in medical decision-making rarely consider the question of whether prediction accuracy will be maintained when the model is faced with new, unseen data [1–3]. In addition, testing these AI models at the local level before adoption is costly and therefore may not be feasible [4]. Therefore, AI system improvement in generalizability to increase accuracy and cost-effectiveness is critical for neural prediction in real-world applications.

Medical AI applications must maintain prediction accuracy when encountering heterogeneous data from differing populations. However, population heterogeneity, variations in data collection, and differences in assessment tools pose significant challenges for out-of-distribution (OOD) generalization, largely due to shifts in underlying feature and label distributions. To mitigate these domain shifts, some researchers advocate for developing foundation models trained on large, diverse datasets, which can then be fine-tuned for domain-specific tasks [4–6]. While foundation models have demonstrated remarkable success in natural language processing, rigorously applying them to medical images remains challenging due to inherent biological variances and the continuous, low-entropy nature of visual data [7–9]. In clinical practice, only a small amount of labeled data and a much larger amount of unlabeled data is typically available for reasons such as limitations in data collection [10, 11]. Consequently, it is critical to develop algorithms that can effectively leverage unlabeled data to mitigate biological variances across diverse populations, particularly in pediatric healthcare, where rapid brain development and varying assessment tools introduce additional heterogeneity [12–14].

In contrast, specialized models offer a pragmatic but limited route to generalization in medical applications [7, 15]. Rather than relying on very large datasets, these approaches may select representative samples or features to improve performance under distribution shift. However, aggressive sample filtering can discard clinically meaningful biological variation and may reduce transportability. Domain-adaptation methods, including adversarial and contrastive approaches, instead learn representations that are less sensitive to domain-specific features [16, 17]. Their effectiveness can nevertheless be constrained when stable semantic features are unavailable and when feature shift is accompanied by label shift or limited target-domain labels [18, 19].

Reinforcement learning (RL) moves beyond static adaptation by using feedback to update an agent's policy, offering a potential mechanism for adaptation under changing data distributions. However, RL systems can remain sensitive to modest environmental or distributional changes, which limits generalization in medical and other safety-critical applications [20–23]. Effective exploration is particularly difficult in neural prediction, where labels are scarce and feedback is indirect. Methods that improve exploration while preserving previously learned information are therefore needed for cross-domain neural prediction [24–27].

Directed evolution accelerates Darwinian selection by iteratively generating, screening, and retaining candidate variants with desired properties [28–30]. Inspired by this trial-and-error process, DEM treats candidate target subsets and pseudo-label configurations as variants that are screened, mutated, and selected during adaptation. The framework combines evolutionary search, continual learning, reinforcement learning, and confidence calibration. Pseudo-labels address the absence of target-domain outcome labels, while experience replay and selective reinitialization balance retention of source knowledge against adaptation to new target patterns.

In this study, we apply DEM to neural prediction in children with cochlear implants (CIs) across multiple centers. The cohort provides a challenging test case because heterogeneity spans biological, demographic, clinical, linguistic, and technical factors. The etiology of hearing loss is often unknown, and spoken-language development is more variable in implanted children than in children with normal hearing [31, 32]. Participants also span a broad age range and use English, Spanish, or Cantonese, which may engage partly different language networks [33–35]. Outcomes may additionally vary with rehabilitation practices, family factors, MRI scanners and protocols, surgical techniques, devices, and programming. These sources of variation motivate evaluation under clinically relevant domain shift.

## II. MATERIALS AND METHODS

### *A. Study Design, Participants, and Ethics*

This multi-center retrospective cohort study utilized data collected between July 2009 and March 2022 from three independent clinical centers: Ann & Robert H. Lurie Children's Hospital of Chicago (USA), the Royal Victorian Eye and Ear Hospital (Melbourne, Australia), and the Prince of Wales Hospital (Hong Kong SAR, China). Primary home languages included English (Chicago and Melbourne centers), Spanish (Chicago center), and Cantonese (Hong Kong center) [36]. Written informed consent was obtained from all parents or guardians for the use of neuroimaging and clinical data. Ethical approval was secured from the respective institutional review boards at each site: the Stanley Manne Children's Research Institute, the Royal Children's Hospital Human Research Ethics Committee, and the Joint Chinese University of Hong Kong–New Territories East Cluster Clinical Research Ethics Committee.

A total of 278 pediatric cochlear implant recipients (137 [49.3%] female) were included: 143 Chicago-English, 37 Chicago-Spanish, 81 Melbourne-English, and 17 Hong Kong-Cantonese participants. Of the 450 participants initially enrolled, 172 (38.2%) were excluded because of a primary home language other than English, Spanish, or Cantonese; missing baseline MRI; substantial MRI artifacts; developmental disorders; or gross MRI abnormalities.

### *B. Spoken-Language Outcome Definition*

Spoken-language development was assessed with center-specific, age-appropriate instruments. Chicago used the modified Speech Recognition Index before implantation and at repeated follow-up visits through 36 months. Melbourne used the Peabody Picture Vocabulary Test–4 [37] and Preschool Language Scale–4 or –5 [38] before implantation and at 12, 24, and 36 months. Hong Kong used the LittlEARS Auditory Questionnaire [39, 40] before implantation and at 6, 12, and 24 months. For each center, scores were modeled using a linear mixed-effects model with fixed assessment time and participant-specific random intercepts and slopes. Participant-specific slopes were dichotomized at the center-specific median into higher- and lower-improvement classes. This center-stratified definition reduced differences among instruments while retaining participant-level outcome assignment.

### *C. MRI Acquisition and Preprocessing*

T1-weighted MRI scans obtained before cochlear implantation were processed using Advanced Normalization Tools (ANTs) [41] and resampled to 1-mm isotropic resolution; downstream image preparation was implemented in Python 3.8.19. Deformation-based morphometry characterized whole-brain structural variation [42]. Fifteen central axial slices were extracted from each deformation-based morphometry image, cropped, resized to 128 × 128 pixels, and normalized using ImageNet statistics. All slices inherited the corresponding participant-level outcome label.

In Chicago, scans were performed on 3T Siemens scanners (MAGNETOM Skyra, Vida) using magnetization-prepared rapid gradient-echo (MPRAGE) sequences or on 3T General Electric scanners (DISCOVERY MR750, SIGNA Architect) using 3D brain volume (BRAVO) or Fast Spoiled Gradient Recalled Echo (FSPGR) sequences. Scanning parameters for BRAVO (N=43) were TE = 2.72–3.91 ms, TR = 7.40–9.45 ms, flip angle = 12°, matrix = 512 × 512, 148–512 slices, slice thickness = 1–2 mm, and voxel size = 0.3 × 0.3 × 0.6 mm to 0.5 × 0.5 × 1.4 mm. For FSPGR (N=1), parameters were TE = 4.3 ms, TR = 10.63 ms, flip angle = 20°, matrix = 256 × 256, 101 slices, slice thickness = 1.4 mm, and voxel size = 0.9 × 0.9 × 1.4 mm. For MPRAGE (N=136), parameters included TE = 2.38–3.54 ms, TR = 1490–2200 ms, flip angle = 8–9°, matrix = 192 × 192 to 512 × 512, 108–224 slices, slice thickness = 0.8–1 mm, and voxel size = 0.8 × 0.8 × 0.8 mm to 1 × 1 × 1 mm.

In Melbourne, scans were acquired using 1.5T Siemens scanners (MAGNETOM Area, Avanto, and SymphonyTim) and 3T Siemens scanners (MAGNETOM Trio and Verio) with MPRAGE sequences. Parameters (N=81) included TE = 2.31–4.92 ms, TR = 11–2100 ms, flip angle = 9°–20°, matrix = 576 × 426 to 224 × 198, 142–452 slices, slice thickness = 0.38–0.9 mm, and voxel size = 0.4 × 0.4 × 0.8 mm to 0.9 × 0.9 × 0.9 mm.

In Hong Kong, scans were conducted using 3T Siemens Prisma, General Electric, or Philips Achieva scanners with MPRAGE, BRAVO, or Turbo Field Echo (TFE) sequences. Parameters for the Siemens Prisma scanner were TE = 2.35–2.59 ms, TR = 1800 ms, flip angle = 8°, matrix = 256 × 208 to 640 × 640, 192–320 slices, and slice thickness = 0.69–3 mm. For the General Electric scanner, parameters were TE = 2.68–2.81 ms, TR = 7.62–7.71 ms, flip angle = 12°, matrix = 512 × 512, 146–352 slices, and slice thickness = 1–1.1 mm. For the Philips Achieva scanner, parameters included TE = 3.41–3.59 ms, TR = 7.46–7.77 ms, flip angle = 8°, matrix = 224 × 224 to 224 × 280, 224–250 slices, and slice thickness = 1.1 mm.

### *D. Directed Evolution Framework*

Generalization requires a model to transfer prior knowledge to conditions that differ from its training distribution. Standard reinforcement-learning exploration strategies, including value backups [43] and return-based optimization [44], may be inefficient when uncertainty and data distributions change. DEM therefore frames cross-domain adaptation as an iterative evolutionary search in which candidate target subsets are screened and candidate pseudo-labels are evolved through mutation and selection [45, 46].

DEM combines continual learning (CL) with reinforcement learning (RL). CL supports retention of source-domain representations and reduces catastrophic forgetting [47], whereas RL uses feedback from candidate adaptation states to balance exploitation and exploration [48, 49]. Their combination, referred to here as continual reinforcement learning (CRL), enables sequential target-domain adaptation while preserving useful source-domain knowledge. Catastrophic forgetting in this setting refers to degradation of previously learned source representations, whereas ineffective RL exploration refers to failure to identify improved target-domain states; DEM addresses these distinct problems through frozen source parameters, replay, and policy-guided evolutionary search.

The adaptive CL module used paired source and target columns with joint domain adaptation. Source-column parameters were trained to convergence and then frozen. Target-column parameters were optimized for each target domain while being regularized through the joint alignment objective, limiting interference with source knowledge and facilitating transfer during the screening and evolution phases.

For the RL framework, we integrated evolutionary learning strategies to provide iterative feedback for adaptation. To avoid purely random or ε-greedy exploration in an uncertain environment, we used sample selection, pseudo-label mutation, and beam search to prioritize informative candidate states during training [50, 51].

### *E. Screening and Pseudo-Label Evolution*

DEM includes two learning phases. First, the screening phase identifies high-confidence subsets using confidence calibration designed for domain shift (Fig. 1a, d). The resulting pseudo-labels allow training with unlabeled target-domain adaptation data [52–54]. Second, the evolving phase applies mutation and crossover to diversify candidate pseudo-labels and retain higher-performing candidates (Fig. 1b, c). Together, these phases guide exploration while limiting uncertainty during target-domain adaptation.

Specifically, we first introduced a pseudo-label strategy based on a pretrained source-led model into both selecting and evolving stages to address the issue of label scarcity in the target domain. Then, we iteratively trained the classifier model on a subset of target samples with pseudo-labels during screening learning, where each sample subset (selection pattern) is selected based on log-probabilities derived from a Bernoulli distribution parameterized by probabilities from the policy model. Here, we treated the entire selection pattern as a single action with a joint probability that is the sum of log probabilities of each selection. So RL learning can be formulated as follows:

$$J_{sel}P(a) = \sum_{i=1}^{K} logP(S_i)$$

Where $J_{sel}P(a)$ is the joint probability of the entire selection pattern as one action; $P(S_i)$ is the probability of the $i$-th selection.

Next, candidate pseudo-labels were also updated by joint log-probabilities of Bernoulli distribution parameterized by probabilities from the policy model during evolving learning to enhance the diversity of candidate pseudo-labels.

$$J_{mut}P(a) = \sum_{i=1}^{K} logP(L_i)$$

Where $J_{mut}P(a)$ is the joint probability of the entire mutation pattern as one action; $P(L_i)$ is the probability of the $i$-th mutation (labels).

Because pseudo-label confidence can be biased, we incorporated confidence calibration during training. The reward was based on the performance of a classifier trained on the subset selected by an action. Although the reward cannot be attributed directly to an individual sample, the joint action probability allows the policy to update selection patterns associated with improved adaptation performance. Entropy regularization encouraged exploration, and a patience criterion terminated beam-search branches when accuracy failed to improve for a prespecified number of consecutive iterations.

After several iterations of learning, we collect a batch of $N$ actions to compute the total loss as follows:

$$L = \frac{1}{N}\sum_{n=1}^{N}(-logP(a_n) \cdot R_n - \beta H(a_n))$$

where $P(a_n)$ is the probability of action $a_n$, and $R_n$ is the reward associated with that action. Specifically, the reward is defined as:

$$R_n = L_{Joint}^{(n-1)} - L_{Joint}^{(n)}$$

Where $L_{Joint}^{(n)}$ denotes the joint adaptation loss after executing action $a_n$. This loss is computed using the labelled source data and the pseudo-labelled target adaptation data. Accordingly, $R_n$ quantifies the improvement in adaptation performance induced by the action, with larger reductions in $L_{\text{joint}}$ yielding higher rewards. Ground-truth labels from the target test set are used only for final evaluation and are not involved in policy updates, reward computation, confidence calibration, or model selection.

In addition, $H(a_n)$ is the entropy of the policy for action $a_n$; and $\beta$ is a coefficient that controls the significance of entropy. Here $H(a_n)$ can be defined as:

$$H(a_n) = -\sum_{i=1}^{K_n}\left[p_{n,i} log p_{n,i} + (1-p_{n,i})\log(1-p_{n,i})\right].$$

Where each action $a_n$ consists of $K_n$ independent binary components (e.g., sample selection or label mutation), and $p_{n,i}$ represents the Bernoulli probability for the $i$ component.

DEM is designed to improve exploration efficiency by alternating screening and pseudo-label evolution. The policy is updated from feedback across successive evolutionary-learning iterations, enabling later training actions to reflect the performance of earlier candidate selections. This formulation balances short-term rewards from individual actions with longer-term policy optimization.

### *F. Experience Replay and Continual Reinitialization*

A key challenge in RL is balancing exploitation and exploration. We used experience replay to revisit informative prior states and support stability [55]. Beam search was incorporated into the replay buffer to prioritize promising experiences. Continual backpropagation complemented replay by reinitializing a small fraction of underused units to preserve plasticity [48]. Replay and continual backpropagation therefore served distinct roles: replay retained useful prior states, whereas selective reinitialization supported learning of new target-domain patterns.

### *G. Confidence Calibration*

Random selection and mutation can be inefficient in large candidate spaces. We therefore introduced a confidence-calibration strategy that prioritizes candidates using calibrated prediction confidence. The mechanism combines protection of successful confidence assignments with partial forgetting of unsuccessful assignments, allowing gradual updates while retaining the capacity to explore alternatives [56].

The confidence calibration approach is inspired by the protecting and forgetting mechanism to ensure the accumulation of incremental changes by calibrating confidences of predictions in every iteration. This means a high and calibrated confidences of predictions will be selected by the protecting and forgetting mechanism based on the feedback from each subset. The protecting term preserves successful search by keeping confidences close to their values from the previous iteration. The forgetting term allows the algorithm to adjusting confidences back toward their initial values, ensuring that bad searches can be 'forgotten' and better ones can be explored. We can formulize this regulation methods as follows:

$$C_{I_j}^{updated} = C_j^{prev} + \underbrace{\lambda\left(C_j^{curr} - C_{I_j}^{prev}\right)^2}_{protecting\ term} - \underbrace{(1-\lambda)\left(C_j^{prev} - C_{I_j}^{init}\right)^2}_{forgetting\ term}$$

Here, C denotes sample confidence; I_j identifies a sample in the current subset; M, Z, and N denote the relevant subset or dataset sizes; and λ controls the balance between protection and forgetting. The adaptive value of λ is defined as follows:

$$\lambda = \frac{1}{1+\exp(-k\cdot\Delta_{acc})}$$

Here, k is a scaling factor that controls the steepness of the transition and $\Delta_{acc}$ is the accuracy change.

### *H. Joint Domain-Adaptation Objective*

To address feature shift, the source-led and target-led models used the joint domain-adaptation objective illustrated in Fig. 2 [18]. The objective combined adversarial domain alignment, correlation alignment (CORAL), maximum classifier discrepancy (MCD), and supervised classification. The source-led model was optimized with labeled source data and unlabeled target data; target outcomes remained unavailable. Its target predictions initialized pseudo-labels for DEM. During subsequent screening and evolution, source components remained frozen while the target feature extractor and classifier were updated with continual backpropagation under the same alignment objective.

Here, the domain discriminator loss is defined as:

$$L_D(x_s, x_t, m) = -\mathbb{E}_{x_s}\sim x_s[Log\ D(m_s(x_s))] - \mathbb{E}_t\sim x_t[Log\ (1 - D(m_t(x_t)))]$$

Here, $D(\cdot)$ is the domain discriminator, $m_s(\cdot)$ is the source feature extractor, $m_t(\cdot)$ is the target feature extractor and $x$ is the source or target image.

Next, CORAL loss (second-order alignment) was denoted as:

$$L_C(x_s, x_t, m) = \frac{1}{4d^2}\sum_{i=1}^{d}\sum_{j=1}^{d}(Cov(m_s(x_s)) - Cov(m_t(x_t)))^2$$

Where, $Cov(\cdot)$ represents the computation of covariance matrix.

Meanwhile, discrepancy loss between the predictions of two classifiers is usually:

$$L_P(C_s, C_t; x) = \frac{1}{N}\sum_{i=1}^{N}\left|\sigma(C_s(m_s(x_s)) - \sigma(C_t(m_t(x_t))\right|$$

Here, $C_s(\cdot)$ is the source classifier, $C_t(\cdot)$ is the target classifier, and $\sigma$ is the sigmoid function.

Finally, the classification loss is defined as:

$$L_{cls}^{(n)} = -\mathbb{E}_{(x,y)}\sim D^{(n)}\sum_{k=1}^{K}1_{[k=y]}\,Log\;C_n(m_n(x))$$

Here, $D^{(\cdot)}$ refers to the source or target data distributions.

In summary, the source-led pretrained model learned domain-invariant representations through joint domain-adaptation optimization and generated initial pseudo-labels for the target domain. During screening and evolution, target-column parameters remained close to the frozen source-column parameters, facilitating knowledge transfer while the target-led models iteratively selected target samples and updated pseudo-label variants.

## III. EXPERIMENTS

### *A. Network Initialization and Optimization*

Seven ImageNet-pretrained convolutional backbones—AlexNet, VGG19, ResNet-50, Inception-v3, GoogLeNet, MobileNet, and DenseNet-169—were screened for feature extraction (Table I). MobileNet achieved the highest within-domain accuracy and was selected as the common backbone for all subsequent comparisons. Architecture-specific adaptive pooling was applied where required so that each backbone produced a one-dimensional feature vector before binary classification. Training used binary cross-entropy with logits, Adam optimization (learning rate, $1 \times 10^{-4}$), a batch size of 64 images, and a maximum of 100 epochs. Random rotation and horizontal flipping were applied during training, and early stopping was triggered after 20 epochs without improvement in validation loss.

### *B. Participant-Level Partitioning*

For within-domain experiments, each center-language dataset was divided into 80% development data and a 20% held-out test set. Five-fold cross-validation was performed within the development data for model selection. All 15 MRI slices from a participant were assigned to the same partition and fold; therefore, no participant contributed data to both training and testing. Outcome classes were defined using center-stratified median splits, as described in Section II-B.

### *C. Baseline and Domain-Adaptation Comparisons*

To compare the proposed DEM against representative approaches for cross-domain neural prediction, we evaluated two commonly adopted paradigms: deep transfer learning and domain adaptation. Deep transfer learning represents the conventional strategy for adapting neural prediction models to new datasets. In this study, we evaluated the applicability of deep transfer learning models across four distinct CI datasets under both within-domain and cross-domain settings to gain insights into their generalizability. First, we assessed the performance of models on various single datasets, including children in the Chicago area who were learning English (henceforth “Chicago English”) or Spanish (“Chicago Spanish”), children in the Melbourne area learning English (“Melbourne English”), children in Hong Kong learning Cantonese (“Hong Kong Cantonese”), as well as on their combined datasets: Chicago + Melbourne, and Chicago + Melbourne + Hong Kong. Furthermore, we tested whether the model trained on the largest dataset (Chicago English) could predict outcomes for three specific cross-domain scenarios: 1) learning Spanish at the same center (Chicago Spanish), 2) learning the same language at another medical center (Melbourne English), and 3) learning a different language at a different center (Hong Kong Cantonese).

To evaluate whether conventional representation learning is sufficient to mitigate domain shift, we implemented a domain adaptation framework that aligns feature distributions between labeled source data and unlabeled target data. The framework jointly optimized source classification together with multiple feature-alignment objectives, including adversarial domain alignment, correlation alignment, and maximum classifier discrepancy, enabling invariant feature learning across domains. During adaptation, only source labels and unlabeled target data were available, while target ground-truth labels were reserved exclusively for final evaluation. To evaluate the generalizability of domain adaptation under both feature shifts and label shifts, we assessed whether a transfer learning model with domain adaptation, trained on the largest dataset (Chicago English), could adapt to various OOD scenarios. Specifically, we examined how well the model adapted to CI candidates learning a different language at the same center (Chicago Spanish), learning the same language at a different center (Melbourne English), and learning a different language at a different center (HK Cantonese). All baseline methods employed the same backbone network, training protocol, and target datasets as DEM. The complete optimization objective is provided in Section II-H.

### *D. DEM Cross-Domain Evaluation Protocol*

Chicago English, the largest dataset, served as the labeled source domain. The target domains were Chicago Spanish (same center, different language), Melbourne English (different center, same language), and Hong Kong Cantonese (different center and language). For each target domain, 80% of participants provided unlabeled adaptation data and the remaining 20% formed a held-out labeled test set [57]. A source-led model first generated initial target pseudo-labels; its parameters were then frozen for CRL. Screening selected high-confidence target samples, whereas evolution refined pseudo-label variants. Target test labels were used only for final evaluation and never for optimization, pseudo-label selection, confidence calibration, early stopping, or reward computation. Direct transfer learning, joint domain adaptation, and DEM used identical source and target partitions, backbone architecture, and optimization settings.

### *E. Ablation Study Design*

Ablation experiments in the Melbourne target domain examined three design choices. First, conventional RL, CRL without domain adaptation, and CRL with domain adaptation were compared to isolate the contributions of continual learning and feature alignment. Second, CRL trained from scratch was compared with CRL using continual reinitialization to assess the value of preserving useful representations while restoring plasticity. Third, fixed-threshold screening with random mutation was compared with confidence-calibrated screening and evolution. These ablations were confined to the Melbourne target domain and were not used to support claims about all target domains.

### *F. Evaluation Metrics and Reproducibility*

Performance was assessed using accuracy, sensitivity, specificity, and area under the receiver operating characteristic curve (AUC), with 95% confidence intervals. The held-out target test set was evaluated only after model development was complete. The clinical and neuroimaging data are not publicly available because they contain potentially identifiable information and are governed by institutional approvals and data-use agreements. Deidentified data may be requested from the corresponding author, subject to institutional approval and an appropriate data-use agreement. Code for DEM and the reported analyses is available at https://github.com/DLDLCQJ/DEM.

## IV. RESULTS

### *A. Backbone Selection and Within-Domain Performance*

We first compared seven pretrained convolutional backbones using the Chicago-English dataset to identify the backbone for subsequent experiments (Table I). All architectures achieved reasonable predictive performance, with accuracies ranging from 79.95% to 90.91%. MobileNet achieved the highest overall performance, with an accuracy of 90.91% (95% CI, 89.41%–92.41%) and an AUC of 0.908 (95% CI, 0.893–0.923), followed by DenseNet-169 (accuracy, 89.09%; AUC, 0.890) and ResNet-50 (accuracy, 88.02%; AUC, 0.880). MobileNet was therefore selected as the common backbone for subsequent cross-domain experiments.

We next evaluated MobileNet under within-domain and combined-dataset settings (Table II and Fig. 3). Within-domain performance remained consistently high across Chicago English, Melbourne English, and Chicago Spanish, with accuracies of 89.56%, 90.62%, and 90.81%, respectively, and corresponding AUCs of 0.943, 0.949, and 0.977. Models trained on the combined Chicago–Melbourne and Chicago–Melbourne–Hong Kong datasets achieved accuracies of 86.00% and 86.79%, respectively. These results indicate that the selected backbone achieved stable predictive performance when training and evaluation data were drawn from matched or jointly represented domains.

### *B. Failure of Transfer learning Under Domain Shift*

We subsequently evaluated whether the Chicago-English model generalized directly to unseen target domains (Table II and Fig. 3). In contrast to the strong within-domain performance, direct transfer learning approached chance-level performance across all three target datasets. Accuracy was 50.95% (95% CI, 49.14%–52.75%) for Melbourne English, 50.27% (95% CI, 46.78%–53.76%) for Chicago Spanish, and 50.75% (95% CI, 47.62%–53.87%) for Hong Kong Cantonese, with corresponding AUCs of 0.511, 0.499, and 0.500. This substantial decline relative to within-domain evaluation demonstrates the limited transportability of static transfer learning under center and language shifts.

### *C. Static Domain Adaptation Does Not Fully Resolve the Shift*

Domain adaptation preserved relatively high source-domain performance but failed to achieve reliable discrimination in the held-out target test sets (Table III). Target accuracy and AUC were 43.81% (95% CI, 38.32%–49.31%) and 0.438 for Melbourne English, 55.50% (95% CI, 43.52%–67.74%) and 0.556 for Chicago Spanish, and 53.73% (95% CI, 34.43%–73.02%) and 0.545 for Hong Kong Cantonese, respectively. In contrast, source-domain accuracy remained 84.09%–86.28% across these experiments. This source–target disparity shows that conventional feature alignment alone was insufficient to overcome the evaluated shifts.

### *D. DEM Improves Cross-Domain Target Performance*

DEM consistently improved target-domain performance (Table IV and Fig. 4). From the screening phase to the evolving phase, target accuracy increased from 62.71% to 70.73% for Melbourne English, from 65.94% to 79.06% for Chicago Spanish, and from 80.00% to 83.13% for Hong Kong Cantonese; the corresponding evolving-phase AUCs were 0.692, 0.779, and 0.833. Relative to direct transfer learning, the final DEM accuracy was higher by 19.8 percentage points for Melbourne English, 28.8 percentage points for Chicago Spanish, and 32.4 percentage points for Hong Kong Cantonese. The gains across center shift, language shift, and combined center-and-language shift indicate that iterative model evolution recovered target-relevant structure that neither static transfer nor conventional domain adaptation captured. These experiments used unlabeled target-domain data during adaptation and therefore evaluate unsupervised domain adaptation rather than deployment to a completely untouched cohort.

### *E. Ablation Analysis: Why DEM Works*

Ablation analyses in the Melbourne target domain further examined the contribution of individual DEM components (Fig. 5). First, both continual reinforcement learning (CRL) variants outperformed conventional reinforcement learning, and CRL with domain adaptation outperformed CRL without domain adaptation (Fig. 5a), supporting the complementary value of continual optimization and feature alignment. Second, continual reinitialization outperformed training each candidate from scratch (Fig. 5b), indicating that selective reinitialization improved model plasticity during adaptation. Third, confidence-calibrated screening and evolution outperformed fixed-threshold screening with random mutation (Fig. 5c), supporting confidence-guided exploration of target samples and pseudo-label configurations. Taken together, the ablation results attribute DEM's improvement to the interaction of domain alignment, continual reinitialization, and confidence-guided evolution.

## V. DISCUSSION

Domain shift and label scarcity remain major barriers to clinical use of medical AI [4, 10, 20]. To address this challenge, we propose DEM, a novel evolutionary computation framework applied to the forecasting of spoken language improvement. Our comprehensive multi-center evaluation demonstrates that DEM effectively adapts to different centers and language environments, in addition to variability inherent to the child, to forecast spoken language improvements over a period of up to three years for children who will receive CIs. Moreover, ablation experiments further supported the contribution of domain alignment, continual reinitialization, and confidence calibration, suggesting that these components collectively enhance model adaptability under distribution shifts. Importantly, grounded in the principle of biological directed evolution, DEM exhibits robust OOD generalization, indicating its potential for broad application across diverse fields, including but not limited to neural prediction, precision education, drug development, and autonomous systems.

Cross-domain strategies range from large foundation models to specialized adaptation methods [4–7, 15, 20]. DEM follows the specialized approach but extend conventional one-time representation alignment through iterative sample selection and pseudo-label evolution, enabling the model to progressively adapt to distributional shifts using unlabeled target-domain data. This strategy is particularly relevant to clinical settings where labeled outcomes are scarce or unavailable, but unlabeled target samples can be accessed before deployment. Our findings demonstrate the potential of this adaptive framework to improve cross-domain generalization in predicting language improvement following cochlear implantation. Collectively, these findings support a transition from static to adaptive representation learning, providing a potential framework for developing medical AI systems that can maintain robust predictive performance across evolving clinical settings.

DEM's performance depends on exploration efficiency [58]. Confidence-guided screening prioritizes more reliable target samples, while pseudo-label evolution explores alternative label assignments to identify potentially improved adaptation states. Meanwhile, experience replay preserves useful previously acquired information, and continual reinitialization maintains model plasticity by renewing underutilized units. Together, these mechanisms address complementary aspects of adaptation: source-knowledge retention, target-state exploration, and control of pseudo-label uncertainty. This coordinated adaptation may help mitigate the progressive accumulation of erroneous pseudo-labels while preserving the capacity to learn from unseen target distributions.

Several limitations should be considered. First, although DEM demonstrates improved OOD performance relative to the evaluated baselines in this multi-center prognostic cohort, prospective validation in larger, more temporally and geographically diverse populations is required to establish broader clinical generalizability. Second, while the center-stratified binary outcome improved comparability across site-specific assessment instruments, it inherently discards fine-grained information present in continuous developmental trajectories. Third, the current reward is derived from internal adaptation feedback and may therefore be sensitive to inaccurate or overconfident pseudo-labels under severe distribution shifts. Although the memory-based confidence mechanism was designed to regulate iterative adaptation, probabilistic calibration was not explicitly evaluated. Finally, while DEM learns adaptive representations across domains, their biological and clinical interpretability remains limited. Future studies should focus on explicit learning of semantic representations grounded in clinically and biologically meaningful mechanisms to further improve both the interpretability and robustness of adaptive medical AI.

## VI. CONCLUSION

DEM provides a novel adaptive learning framework for neural prediction under clinically relevant center and language shifts. By combining joint domain adaptation with continual reinforcement screening, pseudo-label evolution, replay, and confidence calibration, DEM improved held-out target performance relative to direct transfer learning and conventional alignment baselines in this retrospective cochlear-implant cohort.

## ACKNOWLEDGMENT

Datasets used in this study to illustrate the usage of DEM were collected in conjunction with two projects funded by the US National Institutes of Health (R21DC016069 and R01DC019387) and a project funded by the Research Grants Council of Hong Kong (GRF14605119). These projects do not concern DEM. We thank Shani Dettman and Dawn Choo for providing the dataset from Melbourne.

## REFERENCES

[1] Chekroud, A. M., Hawrilenko, M., Loho, H., Bondar, J., Gueorguieva, R., Hasan, A., Kambeitz, J., Corlett, P. R., Koutsouleris, N., Krumholz, H. M., Krystal, J. H., & Paulus, M. (2024). Illusory generalizability of clinical prediction models. Science, 383(6679), 164-167. https://doi.org/10.1126/science.adg8538

[2] Poldrack, R. A., Huckins, G., & Varoquaux, G. (2020). Establishment of best practices for evidence for prediction: a review. JAMA Psychiatry, 77(5), 534-540.

[3] Wu, J., Li, J., Eickhoff, S. B., Scheinost, D., & Genon, S. (2023). The challenges and prospects of brain-based prediction of behaviour. Nature Human Behaviour, 7(8), 1255-1264. https://doi.org/10.1038/s41562-023-01670-1

[4] Lenharo, M. (2024). The testing of AI in medicine is a mess. Here's how it should be done. Nature, 632(8026), 722-724.

[5] Ash, J., & Adams, R. P. (2020). On warm-starting neural network training. Advances in neural information processing systems, 33, 3884-3894.

[6] Hager, P., Jungmann, F., Holland, R., Bhagat, K., Hubrecht, I., Knauer, M., Vielhauer, J., Makowski, M., Braren, R., Kaissis, G., & Rueckert, D. (2024). Evaluation and mitigation of the limitations of large language models in clinical decision-making. Nature medicine, 30(9), 2613-2622. https://doi.org/10.1038/s41591-024-03097-1

[7] Goetz, L., Seedat, N., Vandersluis, R., & van der Schaar, M. (2024). Generalization—a key challenge for responsible AI in patient-facing clinical applications. NPJ digital medicine, 7(1), 126.

[8] Moor, M., Banerjee, O., Abad, Z. S. H., Krumholz, H. M., Leskovec, J., Topol, E. J., & Rajpurkar, P. (2023). Foundation models for generalist medical artificial intelligence. Nature, 616(7956), 259-265.

[9] Rajpurkar, P., Chen, E., Banerjee, O., & Topol, E. J. (2022). AI in health and medicine. Nature medicine, 28(1), 31-38.

[10] Castro, D. C., Walker, I., & Glocker, B. (2020). Causality matters in medical imaging. Nature communications, 11(1), 3673. https://doi.org/10.1038/s41467-020-17478-w

[11] Yu, G., Sun, K., Xu, C., Shi, X.-H., Wu, C., Xie, T., Meng, R.-Q., Meng, X.-H., Wang, K.-S., Xiao, H.-M., & Deng, H.-W. (2021). Accurate recognition of colorectal cancer with semi-supervised deep learning on pathological images. Nature communications, 12(1), 6311. https://doi.org/10.1038/s41467-021-26643-8

[12] Bethlehem, R. A. I., Seidlitz, J., White, S. R., Vogel, J. W., Anderson, K. M., Adamson, C., Adler, S., Alexopoulos, G. S., Anagnostou, E., Areces-Gonzalez, A., Astle, D. E., Auyeung, B., Ayub, M., Bae, J., Ball, G., Baron-Cohen, S., Beare, R., Bedford, S. A., Benegal, V., . . . Vetsa. (2022). Brain charts for the human lifespan. Nature, 604(7906), 525-533. https://doi.org/10.1038/s41586-022-04554-y

[13] Courchesne, E., Pierce, K., Schumann, C. M., Redcay, E., Buckwalter, J. A., Kennedy, D. P., & Morgan, J. (2007). Mapping early brain development in autism. Neuron, 56(2), 399-413. https://doi.org/10.1016/j.neuron.2007.10.016

[14] Guan, H., & Liu, M. (2022). Domain Adaptation for Medical Image Analysis: A Survey. IEEE Trans Biomed Eng, 69(3), 1173-1185. https://doi.org/10.1109/tbme.2021.3117407

[15] Li, M. M., Reis, B. Y., Rodman, A., Cai, T., Dagan, N., Balicer, R. D., Loscalzo, J., Kohane, I. S., & Zitnik, M. (2026). Scaling medical AI across clinical contexts. Nat Med, 32(2), 439-448. https://doi.org/10.1038/s41591-025-04184-7

[16] Tan, C., Sun, F., Kong, T., Zhang, W., Yang, C., & Liu, C. (2018). A survey on deep transfer learning. Artificial Neural Networks and Machine Learning–ICANN 2018: 27th International Conference on Artificial Neural Networks, Rhodes, Greece, October 4-7, 2018, Proceedings, Part III 27,

[17] Wang, J., Lan, C., Liu, C., Ouyang, Y., Qin, T., Lu, W., Chen, Y., Zeng, W., & Philip, S. Y. (2022). Generalizing to unseen domains: A survey on domain generalization. IEEE transactions on knowledge and data engineering, 35(8), 8052-8072.

[18] Ganin, Y., Ustinova, E., Ajakan, H., Germain, P., Larochelle, H., Laviolette, F., March, M., & Lempitsky, V. (2016). Domain-adversarial training of neural networks. Journal of machine learning research, 17(59), 1-35.

[19] Kouw, W. M., & Loog, M. (2019). A review of domain adaptation without target labels. IEEE transactions on pattern analysis and machine intelligence, 43(3), 766-785.

[20] Futoma, J., Simons, M., Panch, T., Doshi-Velez, F., & Celi, L. A. (2020). The myth of generalisability in clinical research and machine learning in health care. Lancet Digit Health, 2(9), e489-e492. https://doi.org/10.1016/s2589-7500(20)30186-2

[21] Lee, C. S., & Lee, A. Y. (2020). Clinical applications of continual learning machine learning. The Lancet Digital Health, 2(6), e279-e281.

[22] Packer, C., Gao, K., Kos, J., Krähenbühl, P., Koltun, V., & Song, D. (2018). Assessing generalization in deep reinforcement learning. arXiv preprint arXiv:1810.12282.

[23] Parisi, G. I., Kemker, R., Part, J. L., Kanan, C., & Wermter, S. (2019). Continual lifelong learning with neural networks: A review. Neural Networks, 113, 54-71.

[24] Sutton, R. S. (2015). Introduction to reinforcement learning with function approximation. Tutorial at the conference on neural information processing systems,

[25] Taylor, M. E., & Stone, P. (2009). Transfer learning for reinforcement learning domains: A survey. Journal of machine learning research, 10(7).

[26] Cockburn, J., Man, V., Cunningham, W. A., & O'Doherty, J. P. (2022). Novelty and uncertainty regulate the balance between exploration and exploitation through distinct mechanisms in the human brain. Neuron, 110(16), 2691-2702. e2698.

[27] Jiang, Y., Kolter, J. Z., & Raileanu, R. (2024). On the importance of exploration for generalization in reinforcement learning. Advances in neural information processing systems, 36.

[28] Arnold, F. H. (1998). Design by directed evolution. Accounts of chemical research, 31(3), 125-131.

[29] Arnold, F. H. (2017). Directed evolution: bringing new chemistry to life. Angewandte Chemie (International Ed. in English), 57(16), 4143.

[30] Miikkulainen, R., & Forrest, S. (2021). A biological perspective on evolutionary computation. Nature Machine Intelligence, 3(1), 9-15.

[31] Feng, G., Ingvalson, E. M., Grieco-Calub, T. M., Roberts, M. Y., Ryan, M. E., Birmingham, P., Burrowes, D., Young, N. M., & Wong, P. C. (2018). Neural preservation underlies speech improvement from auditory deprivation in young cochlear implant recipients. Proceedings of the National Academy of Sciences, 115(5), E1022-E1031.

[32] Yuan, D., Chang, W. T., Ng, I. H.-Y., Tong, M. C., Chu, W. C., Young, N. M., & Wong, P. C. (2025). Predicting Auditory Skill Outcomes After Pediatric Cochlear Implantation Using Preoperative Brain Imaging. American Journal of Audiology, 34(1), 51-59.

[33] Bialystok, E. (2017). The bilingual adaptation: How minds accommodate experience. Psychol Bull, 143(3), 233-262. https://doi.org/10.1037/bul0000099

[34] Ge, J., Peng, G., Lyu, B., Wang, Y., Zhuo, Y., Niu, Z., Tan, L. H., Leff, A. P., & Gao, J. H. (2015). Cross-language differences in the brain network subserving intelligible speech. Proc Natl Acad Sci U S A, 112(10), 2972-2977. https://doi.org/10.1073/pnas.1416000112

[35] Niparko, J. K., Tobey, E. A., Thal, D. J., Eisenberg, L. S., Wang, N.-Y., Quittner, A. L., Fink, N. E., & Team, C. I. (2010). Spoken language development in children following cochlear implantation. Jama, 303(15), 1498-1506.

[36] Wang, Y., Yuan, D., Dettman, S., Choo, D., Xu, E. S., Thomas, D., Ryan, M. E., Wong, P. C. M., & Young, N. M. (2026). Forecasting Spoken Language Development in Children With Cochlear Implants Using Preimplant Magnetic Resonance Imaging. JAMA Otolaryngol Head Neck Surg, 152(3), 232-241. https://doi.org/10.1001/jamaoto.2025.4694

[37] L. M. Dunn and D. M. Dunn, Peabody Picture Vocabulary Test, 4th ed. Minneapolis, MN, USA: Pearson, 2007.

[38] I. L. Zimmerman, V. G. Steiner, and R. E. Pond, Preschool Language Scale, 5th ed. San Antonio, TX, USA: Pearson, 2011.

[39] M. P. Bagatto, C. L. Brown, S. T. Moodie, and S. D. Scollie, “External validation of the LittlEARS Auditory Questionnaire with English-speaking families of Canadian children with normal hearing,” Int. J. Pediatr. Otorhinolaryngol., vol. 75, no. 6, pp. 815–817, 2011.

[40] H. Liu et al., “Assessment and monitoring of the LittlEARS Auditory Questionnaire used for young hearing-aid users in auditory speech development,” J. Audiol. Speech Pathol., pp. 291–294, 2015.

[41] Tustison, N. J., Cook, P. A., Klein, A., Song, G., Das, S. R., Duda, J. T., Kandel, B. M., van Strien, N., Stone, J. R., Gee, J. C., & Avants, B. B. (2014). Large-scale evaluation of ANTs and FreeSurfer cortical thickness measurements. Neuroimage, 99, 166-179. https://doi.org/10.1016/j.neuroimage.2014.05.044

[42] Gaser, C., Nenadic, I., Buchsbaum, B. R., Hazlett, E. A., & Buchsbaum, M. S. (2001). Deformation-based morphometry and its relation to conventional volumetry of brain lateral ventricles in MRI. Neuroimage, 13(6), 1140-1145.

[43] Mnih, V., Kavukcuoglu, K., Silver, D., Rusu, A. A., Veness, J., Bellemare, M. G., Graves, A., Riedmiller, M., Fidjeland, A. K., Ostrovski, G., Petersen, S., Beattie, C., Sadik, A., Antonoglou, I., King, H., Kumaran, D., Wierstra, D., Legg, S., & Hassabis, D. (2015). Human-level control through deep reinforcement learning. Nature, 518(7540), 529-533. https://doi.org/10.1038/nature14236

[44] Schulman, J., Wolski, F., Dhariwal, P., Radford, A., & Klimov, O. (2017). Proximal policy optimization algorithms. arXiv preprint arXiv:1707.06347.

[45] Boder, E. T., Midelfort, K. S., & Wittrup, K. D. (2000). Directed evolution of antibody fragments with monovalent femtomolar antigen-binding affinity. Proceedings of the National Academy of Sciences, 97(20), 10701-10705.

[46] Yokobayashi, Y., Weiss, R., & Arnold, F. H. (2002). Directed evolution of a genetic circuit. Proceedings of the National Academy of Sciences, 99(26), 16587-16591.

[47] Kirkpatrick, J., Pascanu, R., Rabinowitz, N., Veness, J., Desjardins, G., Rusu, A. A., Milan, K., Quan, J., Ramalho, T., & Grabska-Barwinska, A. (2017). Overcoming catastrophic forgetting in neural networks. Proceedings of the National Academy of Sciences, 114(13), 3521-3526.

[48] Dohare, S., Hernandez-Garcia, J. F., Lan, Q., Rahman, P., Mahmood, A. R., & Sutton, R. S. (2024). Loss of plasticity in deep continual learning. Nature, 632(8026), 768-774.

[49] Littman, M. L. (2015). Reinforcement learning improves behaviour from evaluative feedback. Nature, 521(7553), 445-451.

[50] Guo, Z. D., & Brunskill, E. (2019). Directed exploration for reinforcement learning. arXiv preprint arXiv:1906.07805.

[51] Wang, Y., Hong, J., Cheraghian, A., Rahman, S., Ahmedt-Aristizabal, D., Petersson, L., & Harandi, M. (2024). Continual test-time domain adaptation via dynamic sample selection. Proceedings of the IEEE/CVF Winter Conference on Applications of Computer Vision,

[52] Arazo, E., Ortego, D., Albert, P., O’Connor, N. E., & McGuinness, K. (2020). Pseudo-labeling and confirmation bias in deep semi-supervised learning. 2020 International joint conference on neural networks (IJCNN),

[53] Cascante-Bonilla, P., Tan, F., Qi, Y., & Ordonez, V. (2021). Curriculum labeling: Revisiting pseudo-labeling for semi-supervised learning. Proceedings of the AAAI conference on artificial intelligence,

[54] Lee, D.-H. (2013). Pseudo-label: The simple and efficient semi-supervised learning method for deep neural networks. Workshop on challenges in representation learning, ICML,

[55] Schaul, T. (2015). Prioritized Experience Replay. arXiv preprint arXiv:1511.05952.

[56] Wang, L., Zhang, X., Li, Q., Zhang, M., Su, H., Zhu, J., & Zhong, Y. (2023). Incorporating neuro-inspired adaptability for continual learning in artificial intelligence. Nature Machine Intelligence, 5(12), 1356-1368.

[57] Bai, P., Miljković, F., John, B., & Lu, H. (2023). Interpretable bilinear attention network with domain adaptation improves drug–target prediction. Nature Machine Intelligence, 5(2), 126-136.

[58] Ladosz, P., Weng, L., Kim, M., & Oh, H. (2022). Exploration in deep reinforcement learning: A survey. Information Fusion, 85, 1-22.

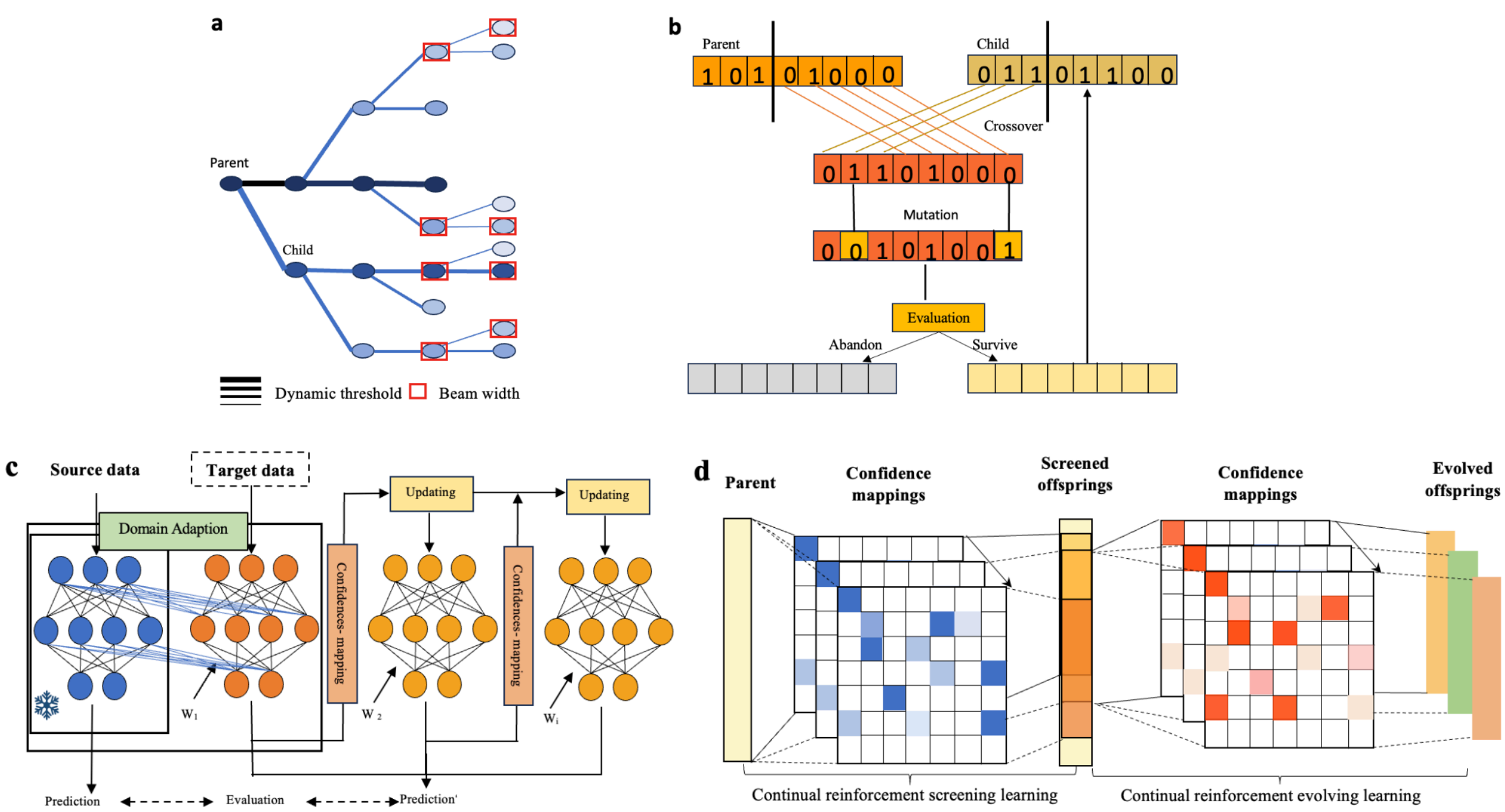

**Fig. 1.** Overview of the Directed Evolution Model (DEM). (a) Beam search is integrated with the replay buffer in continual reinforcement learning to provide structured, prioritized retrieval of informative prior states. (b) Mutation and crossover are combined with beam search to generate and select candidate pseudo-labels and models. (c) In contrast to conventional domain adaptation, which primarily aligns source and target distributions, DEM constructs a dynamic target-data environment through evolutionary screening and pseudo-label evolution, thereby supporting agent–environment interaction without labeled target-domain training data. Continual learning transfers knowledge from the source to the target domain and mitigates catastrophic forgetting. (d) DEM coordinates parallel screening and evolution learners through confidence calibration within a continual reinforcement learning framework.

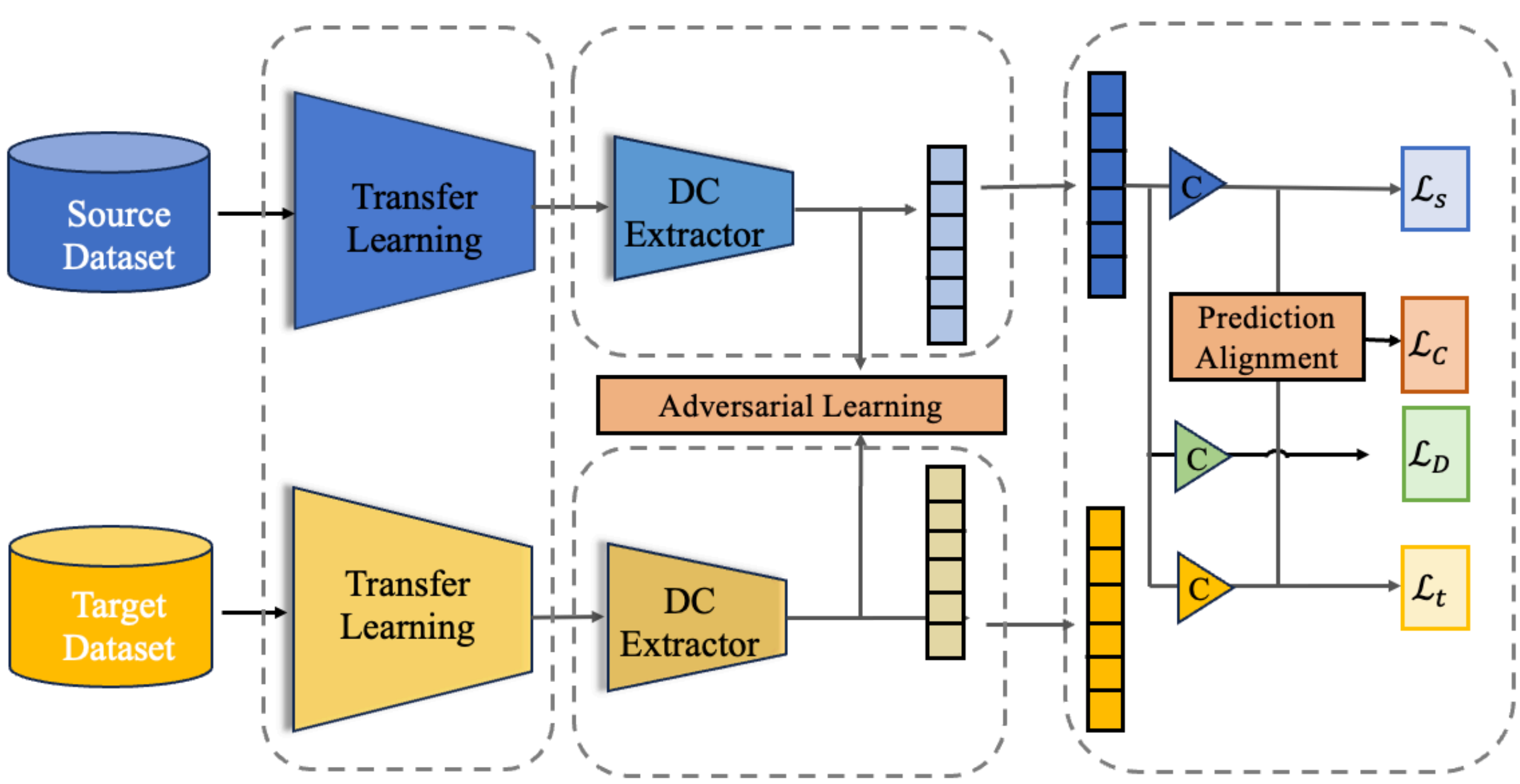

**Fig. 2.** Joint domain-adaptation framework used by the source-led and target-led models. The objective combines adversarial domain alignment, correlation alignment (CORAL), maximum classifier discrepancy, and classification loss; the resulting source-led model initializes pseudo-labels for subsequent DEM screening and evolution.

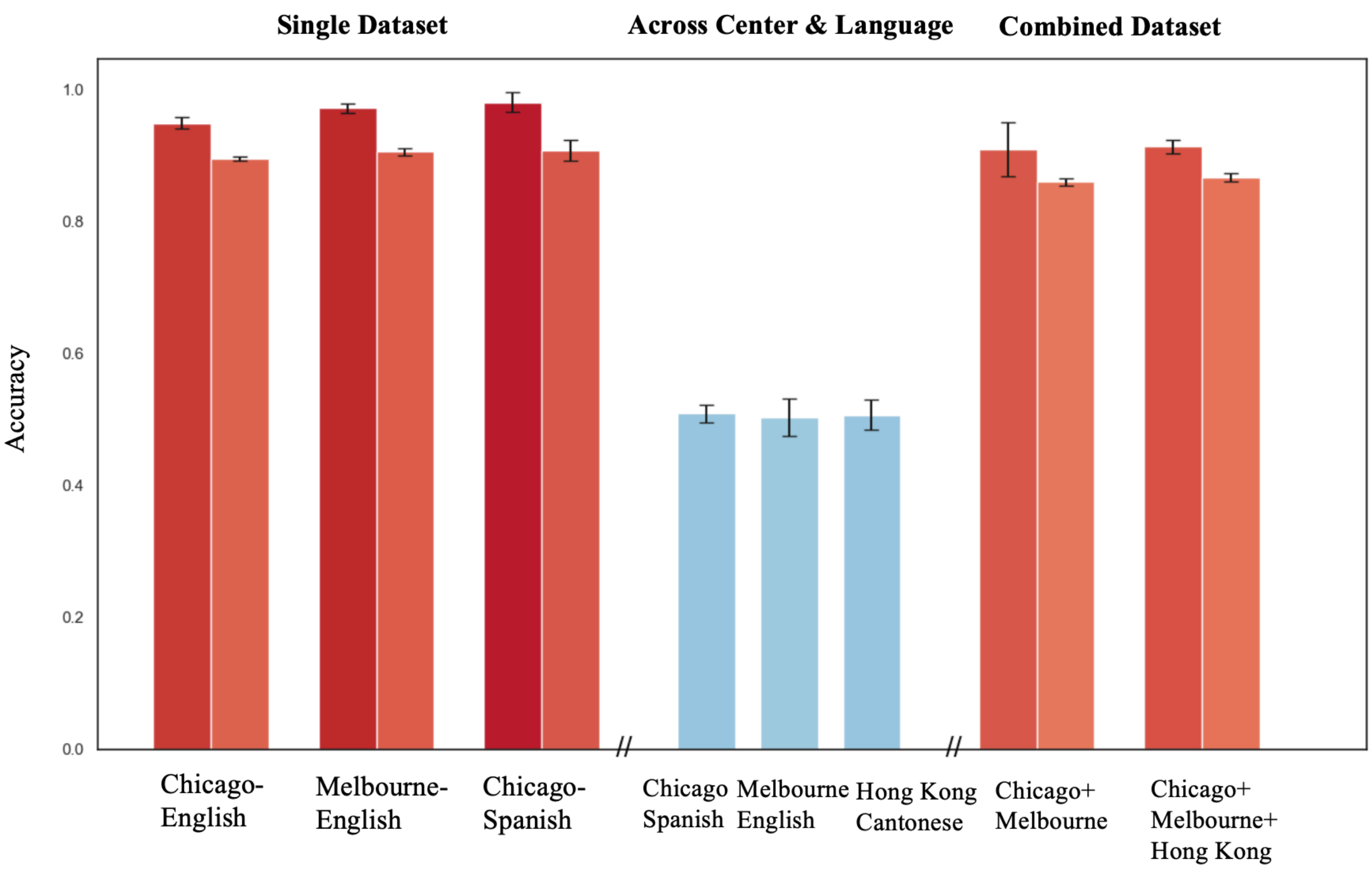

**Fig. 3.** Performance of deep transfer learning under within-domain and cross-domain target evaluation. Models were evaluated using single-center, cross-center or cross-language, and combined datasets. Accuracy remained high during within-domain evaluation (red) but declined on target centers or languages (blue), demonstrating limited cross-domain generalizability of static transfer-learning models.

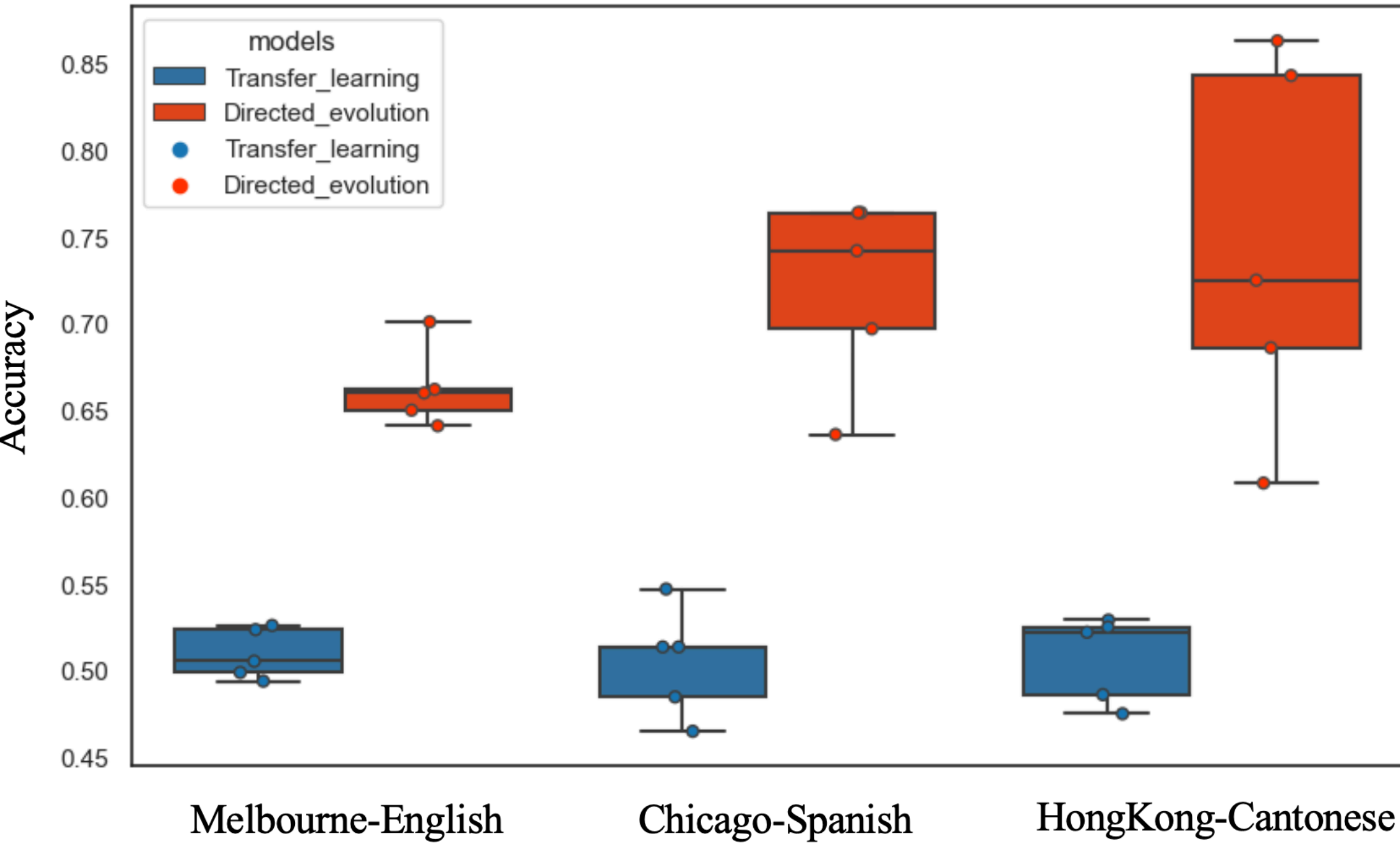

**Fig. 4.** Cross-domain performance of deep transfer learning and the Directed Evolution Model (DEM) across three independent target datasets. The transfer-learning model was trained on labeled source-domain data and directly evaluated on each target-domain test set. DEM was trained on the source domain and then adapted using unlabeled target-domain training data before evaluation on the corresponding held-out target-domain test set. Box plots show the distribution of accuracy across five cross-validation folds.

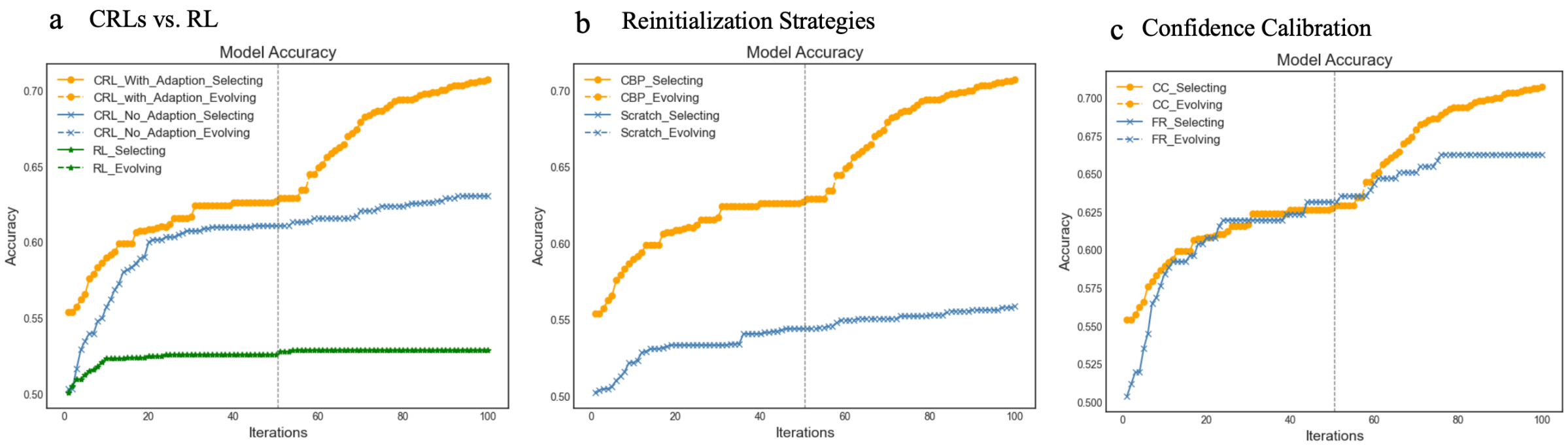

**Fig. 5.** Ablation analysis of key components in the screening and evolution phases of the Directed Evolution Model (DEM). (a) Comparison of learning frameworks. RL denotes domain-specific reinforcement learning; CRL without domain adaptation denotes continual reinforcement learning without feature alignment; and DEM denotes continual reinforcement learning with joint domain adaptation. (b) Comparison of reinitialization strategies. Scratch indicates retraining from an initialized model state, whereas CBP indicates continual backpropagation. (c) Comparison of confidence strategies. RF uses fixed confidence filtering during screening and random mutation during evolution; CC uses the proposed confidence-calibration mechanism in both phases.

**Table I**

Comparative performance within-domain backbone in the Chicago-English group

| Models | % (95% CI) | | | |
|---|---|---|---|---|
| | Accuracy | Sensitivity | Specificity | AUC (95% CI) |
| VGG19 | 81.17 (80.11-82.22) | 86.19 (84.80-87.57) | 75.73 (73.55-77.90) | 0.810 (0.799-0.820) |
| ResNet-50 | 88.02 (86.92-89.11) | 88.16 (85.98-90.34) | 87.86 (86.21-89.51) | 0.880 (0.869-0.891) |
| DenseNet-169 | 89.09 (88.06-90.12) | 92.11 (91.47-92.74) | 85.83 (83.64-88.02) | 0.890 (0.879-0.900) |
| AlexNet | 79.95 (78.61-81.30) | 84.13 (82.67-85.58) | 75.44 (72.53-78.35) | 0.800 (0.786-0.813) |
| Inception-v3 | 83.64 (81.75-85.53) | 85.65 (77.40-93.90) | 81.46 (73.24-89.67) | 0.836 (0.817-0.854) |
| GoogleNet | 87.13 (85.54-88.72) | 92.38 (90.53-94.22) | 81.46 (79.07-83.84) | 0.869 (0.853-0.885) |
| MobileNet | 90.91 (89.41-92.41) | 92.56 (90.56-94.55) | 89.13 (86.97-91.28) | 0.908 (0.893-0.923) |

Note: CI, confidence interval; AUC, area under the receiver operating characteristic curve.

**Table II**

Comparative Performance of Transfer Learning Models under Within-Domain and Cross-Domain Target Evaluation Settings

| Type | Dataset | % (95% CI) | | | |
|---|---|---|---|---|---|
| | | Accuracy | Sensitivity | Specificity | AUC (95% CI) |
| Within-domain evaluation | Chicago English | 90.91 (89.41-92.41) | 92.56 (90.56-94.55) | 89.13 (86.97-91.28) | 0.908 (0.893-0.923) |
| | Melbourne English | 90.62 (89.86-91.38) | 91.83 (90.30-93.37) | 89.43 (88.72-90.14) | 0.949 (0.946-0.953) |
| | Chicago Spanish | 90.81 (88.66-92.96) | 95.41 (93.71-97.11) | 85.20 (81.04-89.36) | 0.977 (0.968-0.986) |
| | Chicago+Melbourne | 86.00 (85.32-86.68) | 81.29 (79.60-82.97) | 90.29 (88.85-91.74) | 0.904 (0.895-0.914) |
| | Chicago+Melbourne+Hong Kong | 86.79 (85.98-87.60) | 89.90 (88.18-91.63) | 83.74 (81.24-86.25) | 0.924 (0.918-0.929) |
| Cross-domain target evaluation | Melbourne English | 50.95 (49.14-52.75) | 62.90 (3.74-100) | 39.28 (0-95.66) | 0.511 (0.489-0.533) |
| | Chicago Spanish | 50.27 (46.78-53.76) | 36.89 (0-93.88) | 63.95 (6.43-100) | 0.499 (0.467-0.532) |
| | Hong Kong Cantonese | 50.75 (47.62-53.87) | 36.67 (0-96.18) | 63.26 (3.46-100) | 0.500 (0.496-0.504) |

Note: CI, confidence interval; AUC, the area under the receiver operating characteristic curve.

**Table III**

Cross-Domain Target Evaluation of Deep Transfer Learning with Domain Adaptation across Source and Target Domains

| Type | Datasets | % (95% CI) | | | |
|---|---|---|---|---|---|
| | | Accuracy | Sensitivity | Specificity | AUC (95% CI) |
| Cross center | Source dataset (Chicago English) | 84.09 (83.32-84.87) | 85.54 (82.70-88.38) | 82.67 (79.65-85.68) | 0.841 (0.833-0.849) |
| | Target dataset (Melbourne English) | 43.81(38.32-49.31) | 46.15 (40.74-51.57) | 41.54(33.75-49.33) | 0.438(0.383-0.493) |
| Cross language | Source dataset (Chicago English) | 85.48 (84.62-86.35) | 85.16 (82.67-87.66) | 85.80 (83.87-87.72) | 0.855 (0.846-0.863) |
| | Target dataset (Chicago Spanish) | 55.50 (43.52-67.74) | 57.25 (40.26-74.25) | 54.00 (45.39-62.61) | 0.556 (0.433-0.677) |
| Cross center & language | Source dataset (Chicago English) | 86.28 (85.40-87.17) | 87.46 (85.80-89.12) | 85.13 (82.46-87.80) | 0.863 (0.854-0.872) |
| | Target dataset (Hong Kong Cantonese) | 53.73 (34.43-73.02) | 62.61 (40.81-84.30) | 46.43 (25.39-67.46) | 0.545 (0.352-0.738) |

Note: CI, confidence interval; AUC, the area under the receiver operating characteristic curve.

**Table IV**

Cross-Domain Target Evaluation of DEM: Comparative Performance on Source and Target Datasets

| Type | Dataset | Phase | % (95% CI) | | | |
|---|---|---|---|---|---|---|
| | | | Accuracy | Sensitivity | Specificity | AUC (95% CI) |
| Cross center | Source dataset | - | 86.88 (84.58-89.17) | 83.94 (73.27-94.61) | 89.46 (82.40-96.52) | 0.869 (0.846-0.892) |
| | Target dataset (Melbourne English) | Screening phase | 62.71(59.12-66.30) | 48.35(43.80-52.90) | 92.24(84.38-100) | 0.603(0.561-0.645) |
| | | Evolving phase | 70.73 (68.42-73.03) | 59.32 (55.13-63.51) | 89.18 (83.43-94.92) | 0.692 (0.669-0.716) |
| Cross language | Source dataset | - | 78.13 (73.79-82.46) | 75.16 (64.33-85.97) | 81.51 (74.54-88.47) | 0.783 (0.742-0.830) |
| | Target dataset (Chicago Spanish) | Screening phase | 65.94 (53.26-78.62) | 55.87 (48.23-63.51) | 91.69 (84.96-98.42) | 0.639 (0.533-0.786) |
| | | Evolving phase | 79.06 (66.07-92.05) | 61.15 (58.70-83.59) | 94.75 (87.06-100) | 0.779 (0.650-0.909) |
| Cross center & language | Source dataset | - | 64.38 (55.99-72.76) | 65.40 (49.87-80.92) | 63.24 (52.82-73.67) | 0.644 (0.560-0.728) |
| | Target dataset (Hong Kong Cantonese) | Screening phase | 80.00 (72.62-87.38) | 83.33 (65.79-100) | 77.04 (64.28-89.80) | 0.802 (0.726-0.878) |
| | | Evolving phase | 83.13 (74.43-91.85) | 86.67 (73.28-100) | 80.00 (65.24-94.76) | 0.833 (0.747-0.920) |

Note: CI, confidence interval; AUC, the area under the receiver operating characteristic curve; DEM, directed evolution model.